\def\year{2022}\relax
\documentclass[letterpaper]{article} % DO NOT CHANGE THIS

\usepackage{caption}
\usepackage{subcaption}

\usepackage{lnfee}  % DO NOT CHANGE THIS
\usepackage{times}  % DO NOT CHANGE THIS
\usepackage{helvet}  % DO NOT CHANGE THIS
\usepackage{courier}  % DO NOT CHANGE THIS
\usepackage[hyphens]{url}  % DO NOT CHANGE THIS
\usepackage{graphicx} % DO NOT CHANGE THIS
\usepackage{natbib}  % DO NOT CHANGE THIS AND DO NOT ADD ANY OPTIONS TO IT
\usepackage{caption} % DO NOT CHANGE THIS AND DO NOT ADD ANY OPTIONS TO IT
\usepackage{comment}
\usepackage{amsmath}
\usepackage{amssymb}
\DeclareCaptionStyle{ruled}{labelfont=normalfont,labelsep=colon,strut=off} % DO NOT CHANGE THIS
\usepackage{algorithm}
\usepackage{algorithmic}

\usepackage{newfloat}
\usepackage{listings}
\floatstyle{ruled}
\newfloat{listing}{tb}{lst}{}
\floatname{listing}{Listing}
\title{DyFEn:\\ Agent-Based Fee Setting in Payment Channel Networks}
\author{
    Kiana Asgari\equalcontrib, Aida Afshar Mohammadian\equalcontrib, Mojtaba Tefagh
}
\affiliations{
    \textsuperscript{\rm}\\
    Sharif University of Technology,\\
    Tehran, Iran\\
    mtefagh@sharif.edu
}

\usepackage{bibentry}
\begin{document}

\maketitle

\begin{abstract}
In recent years, with the development of easy to use learning environments, implementing and reproducible benchmarking of reinforcement learning algorithms has been largely accelerated by utilizing these frameworks. In this article, we introduce the Dynamic Fee learning Environment (DyFEn), an open-source real-world financial network model. It can provide a testbed for evaluating different reinforcement learning techniques. To illustrate the promise of DyFEn, we present a challenging problem which is a simultaneous multi-channel dynamic fee setting for off-chain payment channels. This problem is well-known in the Bitcoin Lightning Network and has no effective solutions. Specifically, we report the empirical results of several commonly used deep reinforcement learning methods on this dynamic fee setting task as a baseline for further experiments. To the best of our knowledge, this work proposes the first virtual learning environment based on a simulation of blockchain and distributed ledger technologies, unlike many others which are based on physics simulations or game platforms.
\end{abstract}

\section{Introduction}

\begin{figure*}
  \centering
  \begin{subfigure}[ct]{0.2\textwidth}
    \centering
    \includegraphics[width=\textwidth]{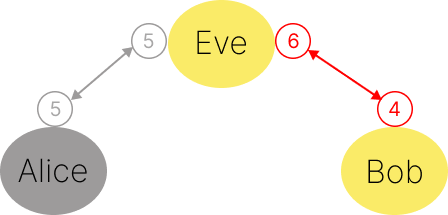}
    \caption{Opening channel}
  \end{subfigure}
  \hfill
  \begin{subfigure}[ct]{0.2\textwidth}
    \centering
    \includegraphics[width=\textwidth]{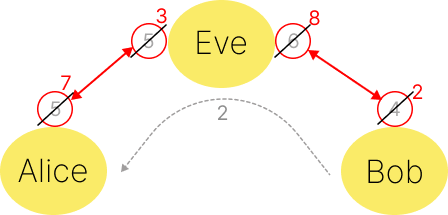}
    \caption{Sending payment}
  \end{subfigure}
  \hfill
  \begin{subfigure}[c]{0.25\textwidth}
    \centering
    \includegraphics[width=\textwidth]{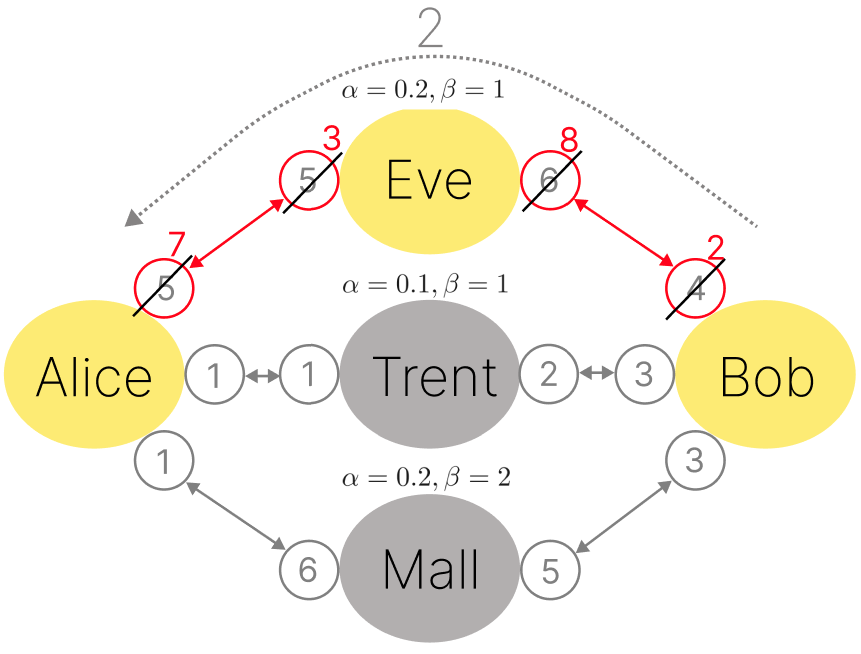}
    \caption{Routing}
  \end{subfigure}
  \begin{subfigure}[ct]{0.25\textwidth}
    \centering
    \includegraphics[width=\textwidth]{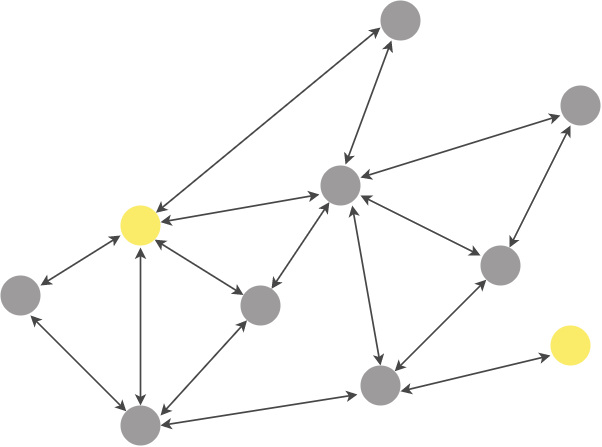}
    \caption{LN graph}
  \end{subfigure}
  \hfill

  \caption{\label{fig:LN}(a) Bob and Eve opening a payment channel. Eve deposits 6 BTC and Bob deposits 4 BTC, resulting in a channel with a capacity of 10 BTC. This transaction is recorded on the Bitcoin blockchain. (b) Bob sends a payment of amount 2 BTC to Alice through Eve. After sending the payment, the balance of each channel used in this transaction will be updated. (c) Finding the best route for sending a 2 BTC payment from Bob to Alice. Following the path through Eve, Bob will pay $\alpha\times amount+\beta = 0.2\times2+1= 1.4$ BTC fee, and following the path through Mall, Bob will pay $\alpha\times amount+\beta=0.2\times2+2= 2.4$ BTC fee. Note that even though the path through Trent needs less fee, there isn't enough liquidity to route the payment. (d) Any two arbitrary nodes in the LN graph can execute a transaction between each other iff there exists a well-funded path between them.}
\end{figure*}

Reinforcement Learning (RL) is a general-purpose formalization of learning through experience. It enables an agent to perform a task by choosing an action and getting a reward signal corresponding to that \cite{sutton2018reinforcement}. RL aims to maximize a notion of cumulative reward based on the feedback to the agent's action and has been able to tackle many problems in the control and planning domains. In the past few years, the advancements in RL were accelerated since the emergence of Deep Reinforcement Learning (DRL), making it possible to use RL in a diverse set of applications.

%%reference to sutton

These advancements have been mainly facilitated by challenging and easy to use environments like Arcade \cite{bellemare2013arcade} and MuJoCo \cite{6386109}. Such environments have played a significant role in the development of the RL algorithms like DDPG \cite{lillicrap2015continuous} and TD3 \cite{fujimoto2018addressing}. Even though a variety of learning environments exist, the performance of DRL in the financial world especially blockchain applications has not been well studied.

The story of the Bitcoin blockchain goes back to a paper authored by the pseudonymous Satoshi \citet{nakamoto2008bitcoin}. Briefly explained, a blockchain is a sequence of blocks chained together, with each block containing a record of transactions. Currently, Bitcoin holds the largest market capitalization among all of the existing cryptocurrencies, but it suffers from the scalability problem caused by the limited space of each block. The most popular solution for the bitcoin scalability issue is the Lightning Network (LN), which will be discussed further.

Since the emergence of LN , transaction fee has always been a key component to incentivize people to join the network and provide liquidity to their payment channels. Naturally, this question arises that which fee policy leads to the maximum profit.

We simplify the fee setting problem to a Markov Decision Process (MDP), with fee rates and base fees corresponding to the channels of a selected node as the actions. This allows us to leverage the state-of-the-art RL algorithms to tackle the fee setting problem. The reward function might be sparse depending on the topological position of the node in the network, making it a challenging environment to learn.

To the best of our knowledge, this study is the first work that provides a benchmark environment for RL based on LN concepts. On the other hand, finding the best fee setting policy in LN is yet unanswered and only a few simple heuristics have been proposed to address the problem. In our paper, we also try to solve this problem with DRL methods. 

In this work, we first explain LN and introduce the fee setting problem for payment channels. Then we discuss our main contributions, a Lightning Experimental Virtual Network (LEViN), and a Dynamic Fee learning Environment (DyFEn). Furthermore, we test the performance and capability of DyFEn in solving the fee setting task by testing it with five different RL algorithms. Finally, we provide a comparison between DyFEn and the already existing strategies for the fee setting problem. Being able to surpass other existing strategies, we hope to take a step forward in using RL for real-world finance problems.

\section{Related Work}

The LN white paper \cite{poon2016bitcoin}, discussed the reason why transaction fee exists in the network and compares the LN fee with the blockchain fee from different aspects. \citet{8726489} proposed a different fee policy which fostered the balancing of channels and improved the network's performance in the long term. \citet{8644930} suggested the fee rate to be proportional to the square root of the channel's capacity, in a simple fee structure which only contains the fee rate. 

LN was meant to be decentralized, but by the evolution of the network, nodes with higher degrees known as merchants appeared which led to several topological studies on LN including \citet{seres2020topological} and \citet{martinazzi2020evolving}. Furthermore, \citet{bartolucci2020percolation} modeled the growth of LN as a bond percolation process. \citet{SSRN} investigated the centralization of LN from the game theoretic point of view.

Developing user-friendly and reliable software systems helps both researchers and the members of LN to experiment with their ideas. Accordingly, \citet{CONOSCENTI2021100717}, same as \citet{Beres2021Cryptoeconomic} enhanced the software support for experimenting with different fee policies in LN by introducing easy to use simulators written in well-known programming languages.

Implementing and designing RL environments have had a fair progress in gaming and control domains, referring to Atari 2600 introduced by \citet{bellemare13arcade} and MuJoCo introduced by \citet{6386109}. 
Despite the capable framework of RL and the rapidly growing area of cryptoeconomics, there have been few works in the intersection of these two areas. \citet{su12125161} did a survey on RL in blockchain-enabled IIoT networks and \citet{8726067} combined these two areas for empowering the next-generation wireless networks.

\section{Preliminaries}

%Lightning Network
Initially, scalability was not a concern in the Bitcoin blockchain. However, with the rising popularity of Bitcoin, this issue has become a drawback of the whole system, making transactions slower and more expensive. LN is an off-chain protocol layered on top of the Bitcoin blockchain, mainly proposed for solving the Bitcoin scalability problem. LN allows the users to execute faster and cheaper transactions outside of the blockchain while preserving the security and decentralization properties of Bitcoin. 

Briefly explained, LN is a network of payment channels modeled as a graph. In this graph, the users are represented by vertices and the payment channels are represented by edges. A payment channel is a connection between two members of the network and is a primary component of LN. Each channel has a corresponding capacity, \emph{i.e.}, the total amount of liquidity available in that channel. Additionally, each peer of the channel has its own balance which is a fraction of the channel's capacity. Peers can have unlimited number of payments up to the limit of their balance. 

% fig of LN

Two arbitrary users in LN can initiate a transaction between each other, if there is a well-funded path between them. A well-funded path is a route consisting of consecutive payment channels where each channel's balance is greater than the amount of the transaction. Each node in this path can charge a small amount of fee in exchange for forwarding the transaction. Furthermore, each peer of these payment channels can independently set and change their own fee setting policy. The final cost of a transaction for the sender will be equal to the sum of transaction amount and the fees of all nodes along the path. 

 The fee protocol of LN is the most important element for incentivizing people to lock their money in the network. While the sender is usually trying to minimize the routing cost of its transaction, channels on the route try to maximize their profit by choosing a good fee setting policy. In the recent implementations of LN, there are two components for fee, \emph{i.e.}, the fee rate ($\alpha$) and the base fee ($\beta$). In return for forwarding $m$ BTC, a routing node will gain $\alpha \times m + \beta$ BTC as income (Figure \ref{fig:LN}).

For modeling the problem, we consider an MDP defined by the tuple $(S,A,P,r,\rho_0,\gamma)$, where $S$ is the state space, $A$ is the action space, $P:S\times A \times S \rightarrow [0, 1]$ is the transition probability distribution, $r:S \times A \rightarrow \mathbb{R}$ is the reward function, $\rho_0:S\rightarrow [0,1]$ is the initial state distribution, and $\gamma \in [0,1]$ is the discount factor. RL algorithms usually try to find a policy $\pi: S\times A\rightarrow  [0,1]$ that maximizes the expected return,
$$J(\pi) = E_{s_0\sim \rho_0, s_t \sim P(.|s_{t-1},a_{t-1}), a_t\sim \pi(.|s_t)} \big[\sum_{t=0}^\infty \gamma^t r(s_t,a_t)\big],$$
where $s_t\in S$ and $a_t\in A$ denote the state of the environment and the action taken by the agent at time step $t$, respectively.

\section{Method}

\subsection{LEViN}

For the task of fee setting, we implemented the Lightning Experimental Virtual Network (LEViN), an advanced simulator that is able to generate random transactions, simulate routing, and calculate metrics of any selected node based on the results of its simulation. LEViN is designed based on the lightning network traffic simulator, known as the LN Simulator, which was introduced by \citet{Beres2021Cryptoeconomic}. We have enhanced LN simulator in several aspects regarding modeling the fee dynamics such as relaxing the strong assumption of constant payment amounts, adding the dynamic fee feature, localizing the simulations, and supporting off-chain rebalancing. We described the simulator in full detail in Appendix A.

\textbf{Data:} Primary topological properties of LN, such as nodes, payment channels, channel fee policies, and channel capacities are available in a snapshot of the network \cite{rohrer19lnattack}. Any arbitrary snapshot of LN can be fed into LEViN, if it adheres to a certain straightforward structure. Additional details of the snapshots and their structure are available in Appendix B. 

\textbf{Balance:} Even though the capacities of the channels are accessible through the LN snapshots, the balances of the two peers of a channel are not available to the public. LEViN sets the balances of the two peers, each equal to half of the channel's capacity. It is also possible to set a channel's balances uniformly or even set them to an arbitrary amount. In the Experiments section, we will show that initial balances have a minor impact on a node's optimal income.

\textbf{Generating transactions:} Real data of transactions can not be extracted from the LN snapshots. For the sake of privacy the publicly available data does not include information about the sender, receiver, and the number of transactions. Thereby, LEViN generates random transactions based on the distribution used in the LN Simulator. The sender of each transaction will be uniformly selected, while the receiver is more likely to be a merchant of the network. Any other arbitrary probability measure can be used to select a merchant. See Appendix A for all the details about generating the transactions.

\textbf{Routing: } Different methods have been proposed for the LN routing problem, namely \cite{hoenisch2018aodv}, \cite{prihodko2016flare}, and \cite{wang2019flash}. For routing transactions, we followed the LN Simulator's assumption of choosing the cheapest possible path. For routing each payment, we found the path with the lowest total fee among all other paths with enough liquidity. Further details can be found in Appendix A.

\textbf{Localization: } LEViN is also able to perform the simulation only for a few nodes close to a selected central node. This property is essential considering the significantly huge number of nodes in the LN and the limited resources available for the simulation. Users can specify the localization size as an input to LEViN. More details about the properties of localization size will be discussed in the Experiments section.

\subsection{DyFEn}
We present a Dynamic Fee learning Environment (DyFEn), designed for addressing the problem of fee setting in the LN. DyFEn is proposed to be used as a challenging testbed for continuous control RL methods. It depends on LEViN to simulate many random payments for each interaction with the agent. 

Assume that we want to achieve optimal income for the node $v$ with $k$ payment channels $c_1,\dots,c_k$ connected to it. In the Experiments section, we will illustrate that only a small number of nodes close to $v$ can have a noticeable effect on its income. Since the changes in  topology of the small neighborhood of $v$ can be effectively slow, we assume the fee setting problem to be fairly stationary. This allows us to propose an MDP for modeling the problem as follows:

\textbf{Observation Space:} We define the observation space to be the set of nonnegative $2k$ dimensional vectors. At each time step $t$, we simulate random transactions in the localized network around $v$. In the observation vector $s_t=(b_1^t,\dots,b_k^t,m_1^t,\dots,m_k^t)$, $b_i$ is the balance of the channel $c_i$ and $m_i$ is the accumulated amount of payments routed through $c_i$. Note that, LEViN computes these values when the $t$th round of simulations is completed.

\textbf{Action Space:} We define the action space as the set of all possible fee rates and base fees for the channels of node $v$. At the time step $t$, the agent chooses the action vector
$a_t=(\alpha_1^t,\dots,\alpha_k^t,\beta_1^t,\dots,\beta_k^t)$ in which $\alpha_i^t$ and $\beta_i^t$ are the fee rate and base fee of channel $c_i$, respectively. Later on, based on the $(\alpha_i^t)_{i=1}^k$ and $(\beta_i^t)_{i=1}^k$ the simulation will be performed, leveraging the dynamic fee property of LEViN.

\textbf{Reward Function:} Our goal is to maximize the income of node $v$. With regard to that, the reward function at time step $t$ is defined by the formula
\begin{equation*}
 r(s_t,a_t) = \sum_{i=1}^k \alpha_i^t m_i^t +\beta_i^t n_i^t,
\end{equation*}
in which $n_i^t$ is the number of transactions routed through $c_i$, provided by LEViN at the end of the time step.

We observed that no transaction will pass through $v$ except for a small fraction of possible fee rates and base fees, resulting in a very sparse reward. For overcoming this problem, we limited the possible fee rates to be less than 1000 and possible base fees to be less than 10000 MSat. See Appendix C for the experiments supporting the sparsity claim.

\section{Experiments}

\begin{figure*}
  \centering
  \begin{subfigure}[b]{0.32\textwidth}
    \centering
    \includegraphics[width=\textwidth]{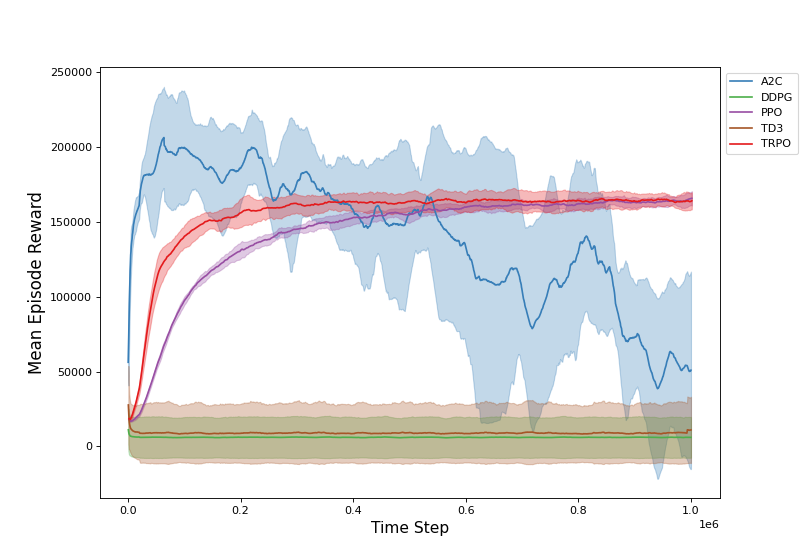}
    \caption{Node $a$}
  \end{subfigure}
  \hspace{0pt}
  \begin{subfigure}[b]{0.32\textwidth}
    \centering
    \includegraphics[width=\textwidth]{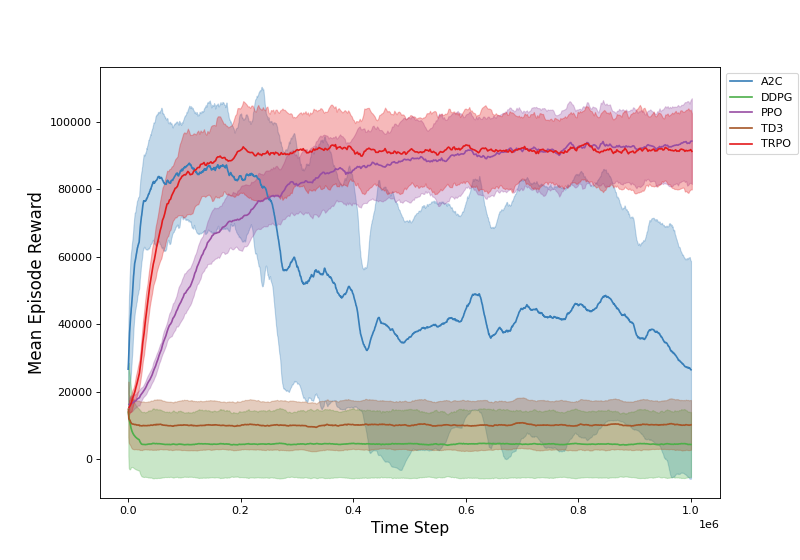}
    \caption{Node $b$}
  \end{subfigure}
  \hspace{0pt}
  \begin{subfigure}[b]{0.32\textwidth}
    \centering
    \includegraphics[width=\textwidth]{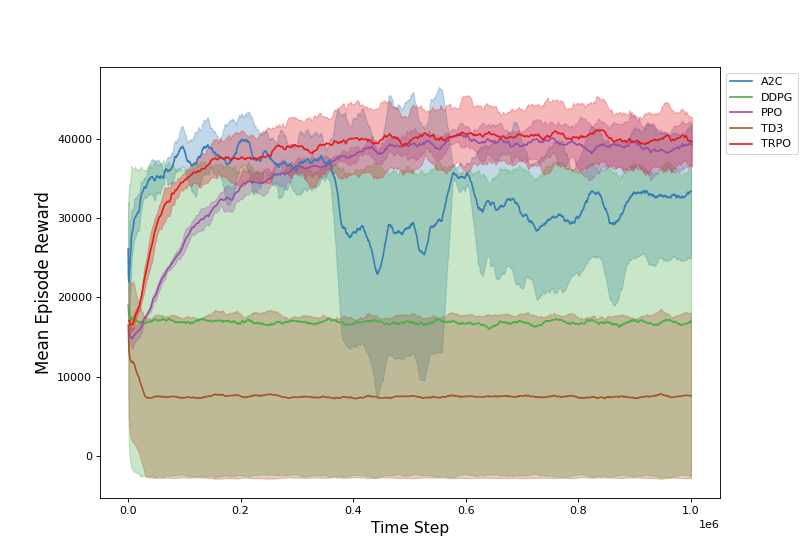}
    \caption{Node $c$}
  \end{subfigure}
  \caption{\label{fig:training}Performance of five well-studied DRL algorithms on three different nodes. The red and purple curves show the convergence of TRPO and PPO, respectively. The green and brown curves are for the off-policy DDPG and TD3 algorithms. Lastly, the blue curve corresponds to A2C algorithm. In addition to the mean episode rewards, the shaded areas depict the standard error of five runs over different random seeds.}

\end{figure*}

We conducted three groups of experiments to answer the following questions:
\begin{enumerate}
    \item Evaluating the performance of state-of-the-art RL algorithms on DyFEn.
    \item How does localizing affect the performance of trained agents?
    \item Whether using RL methods can achieve better incomes compared to the other fee setting baselines. 
\end{enumerate}

To answer (1), we compare the performance of five RL algorithms on three different nodes in LN (Table~\ref{tab:node_table}). To address (2), we train different PPO agents with three localization sizes. We show that smaller localization sizes result in better performances, which highlights the importance of local topological properties of a node in LN. With regard to question (3), our results demonstrate that the trained agents can outperform other baselines by a large margin. Starting from five different initial balances, we also see that RL agents can perform well regardless of the initializations. Next, we elaborate more on each of these questions in the following subsections.

\begin{table}
\begin{center}
\begin{tabular}{c c c c } 
 \hline
 Name & Index & Capacity (Sat)  & Channels per node  \\ [0.5ex] 
 \hline
 $a$ & 97851 & 28154272 & 6 \\ 
 
 $b$ & 71555 &  21498650 & 6 \\

 $c$ & 109618 & 16700000 & 7 \\
 \hline
 LN avg. & &  25800000 &  9  \\
 \hline
\end{tabular}
\caption{\label{tab:node_table}Three randomly chosen nodes with capacity and number of channels close to the average of LN.}
\end{center}
\end{table}

\subsection{Evaluation and Performance Results}

Our main goal in the following experiments is to demonstrate the good behaviors of RL agents while interacting with DyFEn. Being able to gain considerable income requires learning a variety of LN characteristics, such as the different impacts of each channel connected to the node and the trade-off between higher fees and lower amounts of transactions choosing to route through the node. These behaviors will get harder to learn as the number of channels increases which in turn causes the dimension of the action space to grow.

 Figure \ref{fig:training} shows the training curves of five different DRL algorithms. We chose DDPG \cite{lillicrap2015continuous}, TD3 \cite{fujimoto2018addressing}, A2C \cite{mnih2016asynchronous}, PPO \cite{schulman2017proximal}, and TRPO \cite{schulman2015trust} as five well-established representative methods, each trained for 1,000K time steps using five different random seeds. We set the localization size to 100 and considered equal initial balances for both of the peers of each channel in all the experiments. We conducted our experiments based on Stable-baselines3 \cite{stable-baselines3} implementations for both on-policy and off-policy algorithms. We refer the interested readers to Appendix A for the details of the used hyperparameters and the experimental setups.

\begin{figure}[t]
  \centering
        \begin{subfigure}{0.4\textwidth}
        \centering
        \includegraphics[width=\textwidth]{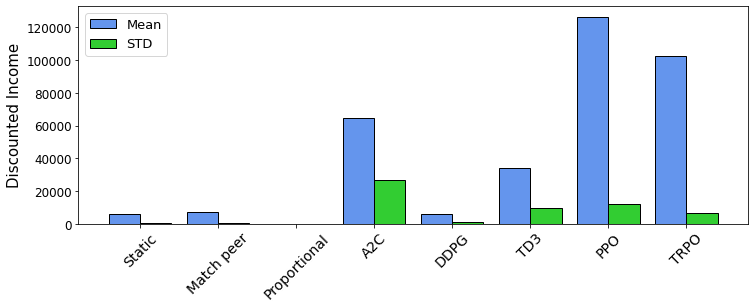}
        \caption{Node a}
        \end{subfigure}
        \begin{subfigure}{0.4\textwidth}
        \centering
        \includegraphics[width=\textwidth]{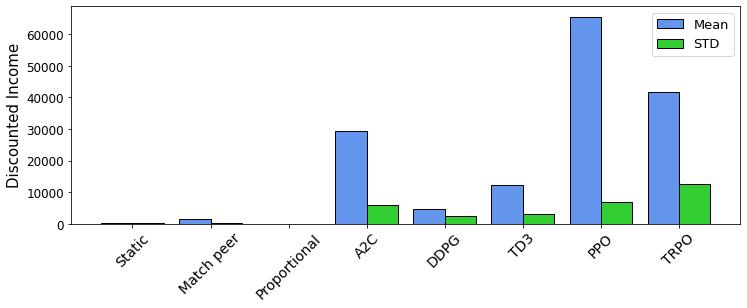}
        \caption{Node b}
        \end{subfigure}   
        \begin{subfigure}{0.4\textwidth}
        \centering
        \includegraphics[width=\textwidth]{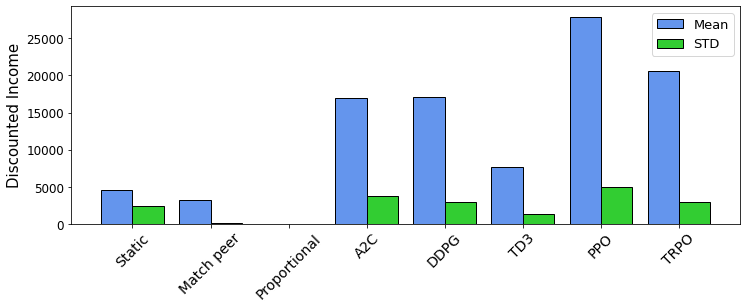}
        \caption{Node c}
        \end{subfigure}%  
    \caption{\label{fig:baseline}Episodic discounted income of three different baselines compared to the trained DRL agents. We used five random initial balances for each strategy. }

\end{figure}

As shown in the training curves, PPO and TRPO performed well in all the experiments, yielding the best solutions in two out of the three nodes. Even though TRPO had faster convergence, we found that trained PPO agents perform better when changing the initial balances (Figure \ref{fig:baseline}). In sharp contrast, TD3 and DDPG performed poorly in all the experiments which confirm the already known difference between training off-policy and on-policy algorithms. Moreover, we found A2C illustrates rather unstable behaviors in comparison to the other on-policy algorithms that showed fast and consistent progress.

\subsection{Sensitivity to Localization Parameter}
Two more tests are conducted on DyFEn to further investigate the properties of LN and the trained agents. To evaluate the importance of the localization size parameter, we trained three PPO models on the node $a$ using localization sizes $L=100,250,500$ with five different random seeds. Afterward, we tested the performance of the trained PPO agents on DyFEn with localization size $L=500$. The results of this group of experiments are summarized in Figure \ref{fig:localizing} (for further details see Appendix C).

Our experiments show that the PPO model trained on DyFEn with a smaller localization size is gaining better incomes in comparison to the PPO model trained on DyFEn with a larger localization size. This demonstrates that the local topological properties of the node are directly affecting its income. Thereby, closer nodes can drastically change one's maximum gain. This highlights the need for a method that is able to simulate LN transactions and predict the maximum income depending on the position of the node. Turning this argument around, this will allow the users to decide on the nodes they want to open a payment channel with in order to optimize their future income.

\begin{figure}
\centering
\includegraphics[width=7cm]{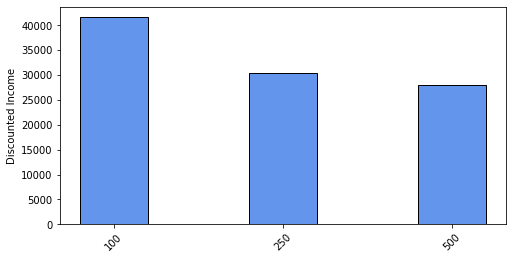}
\caption{\label{fig:localizing} Performance of three trained PPO agents, each trained on $L=100,250,500$. Testing is done using DyFEn with localization size $L=500$.}
\end{figure}

\subsection{Comparison to Heuristic Baselines}
Figure \ref{fig:baseline} illustrates the performance of the trained DRL agents in comparison to three famous heuristic fee setting baselines, namely, static, match peer, and proportional strategies.
\\

\textbf{Static:} In this strategy the node's fee policy is to choose fixed fee rates and base fees during the whole episode. We used the node's real fee rates and base fees from the LN snapshot to conduct our experiments. As Figure \ref{fig:baseline} suggests, current fee policies will result in almost negligible income which is in agreement with the recent observations of the real-world LN. For static fee policies with different base fees and fee rates refer to Appendix C.
\\

\textbf{Match peer:} When using match peer strategy, the fee policies of node's payment channels are determined to match the fee policies of the other peer of each channel. Since the topological properties of two peers of a channel are close, one can expect a good fee policy for one to be also suitable for the other. Right now since the income of most of the users in LN is insignificant, this idea is less likely to work.
\\

\textbf{Proportional:} This strategy is based on the trade-off between channel balance and channel fee rate, \emph{i.e.}, 
\begin{equation*}
    \beta=0,\quad \alpha \sim 1-\frac{\textrm{balance}}{\textrm{capacity}}.
\end{equation*}
The idea behind this strategy is that with a higher fee rate, fewer payments will route through the channel, resulting in more stable balances. On the other hand, when having enough balances, using lower fee rates will result in more payments routing through the channel. We found the proportional strategy to be very sensitive to the initial balances and it results in zero income for most of the initializations. For further details refer to Appendix C.

We used five different random initial balances for conducting these experiment. As illustrated in the plots, one can see a significant improvement when using DRL methods. PPO as our best model can gain almost ten times more income compared to the other strategies with low sensitivity to the initial channel balances.
\\ \\
In conclusion, our results on consistent and fast convergence of PPO (Figure \ref{fig:training}), its stability despite changing the initial balances (Figure \ref{fig:baseline}), and its superior performance while using smaller localization sizes (Figure \ref{fig:localizing}) provide empirical evidence that DRL algorithms are able to solve the fee setting problem better than any other currently proposed heuristic methods, which also demonstrates the possibility of using RL in other real-world financial problems.

\section{Conclusions \& Future Work}

In this work, first we presented LEViN, a highly optimized simulator for the off-chain payment channel networks such as the Bitcoin LN with several novel features targeted towards customizing and evaluating the dynamic fee mechanism. Next, we proposed DyFEn which is built on top of LEViN and provides an environment to learn the difficult but not intractable reference task of finding the optimal fee setting policy. Furthermore, the performance of several well-established DRL algorithms is studied and compared to the heuristics currently used in practice as baseline.

Overall, LEViN in conjunction with DyFEn can provide both a benchmark environment for RL research to investigate the performance of different methodologies and an interface to learn a policy from a snapshot of LN to be used in action. Moreover, on-policy algorithms are the clear winner with a significantly higher income based on our initial experimental results, but further work is needed to implement and test the derived policies in the real-world scenario. In the future, we plan to add more functionalities geared towards other hard tasks which arise in the layer 2 of blockchain protocols.

\section{Acknowledgments}
To be added.
\\ \\ \\ \\
\bibliography{lnfee.bib}

\clearpage
\appendix
\section{Supplementary materials}

In this section, we provide further explanations of our implementations and experiments. 
\begin{itemize}
\item Appendix A: Technical Details.
\item Appendix B: LN Data.
\item Appendix C: Additional Experiments.
\end{itemize}
\section{A. Technical Details}
\subsection{LEViN simulator}
\subsubsection{Generating transactions:} The simulator gets various inputs including a list of transaction types. Each transaction type is specified with three variables, count, amount, and epsilon, referring to the number of transactions, amount of payments, and the fraction of merchants, respectively. Let $n$, $v_i$, and $M$ denote the total number of nodes, node i, and the set of merchant nodes. The sender is uniformly selected while the receiver of each transaction is sampled based on the following distribution:

\begin{align*}
    &\mathbb{P}[\textnormal{receiver}=v_i] =
    \begin{cases}
      0 & \textnormal{sender} = v_i\\
      \frac{\epsilon}{n-1} & v_i \in M \\
      \frac{1-\epsilon}{n-1} & \textnormal{otherwise}
    \end{cases}
\end{align*}

\subsubsection{Routing:} 
We use NetworkX library \cite{SciPyProceedings_11}, written in Python language, for generating weighted graphs and running path searching algorithms on them. The network snapshot graph is a weighted graph $G=(V, E, \alpha, \beta)$ in which $V$ is the set of nodes of the network, $E$ is the set of payment channels, $\alpha: V\times E \longrightarrow \mathbb{R}$ is the corresponding fee rate, and $\beta: V\times E \longrightarrow \mathbb{R}$ is the corresponding base fee to each peer of payment channel. Let $f_{v,e}: \mathbb{R} \longrightarrow \mathbb{R}$ be the fee function which is calculated as,
$$f_{v,e}(a) = \alpha(v,e)\times a + \beta(v,e)$$
where $a$ is the payment amount of the transaction. The simulator generates a weighted bidirected graph $G_a = (V,E,w_a)$ corresponding to the amount of each transaction type where,
$$w_a(e=(u,v)) = f_{u,e}(a)$$

The main speed-up trick behind the simulation is to only keep channels with adequate balance (balance $\geq$ a) in the simulation graph which is a subset of $G_a$ prior to running the Dijkstra shortest path algorithm \cite{dijkstra1959note}. Equivalently, the simulator removes the underbalanced edges (payment channels) from $G_a$ before each round of simulation. This can significantly enhance the overall performance of the simulator.

We hard-coded the assumption that the sender will always choose the cheapest path for routing its transaction. From the implementation point of view, we use the shortest path algorithm implemented in the NetworkX library on the simulation graphs to find the final route for each transaction. 
\\
\subsubsection{Simulation: }
Now that we have the simulation graphs, path searching algorithms, and the set of random transactions, we can initiate each transaction on the corresponding graph and see whether the transaction is successful or not. In LEViN, a transaction can fail in case there is no well-funded path between the sender and the receiver. Note that in LN, the transaction can fail for many other reasons as well, for instance, the middle nodes in the route becoming unresponsive. If the transaction was successful, the balance of active channels gets updated. 

The LEViN simulator gets active channels as an input, which simply indicates the channels which are updated by the end of each round of simulation. Other channels' balances remain unchanged during the entire simulation. This feature of the simulator gives better control over the experiments and helps to reduce the running time. In the experiments we conducted in this paper, the active channels are set to be the channels of the selected node.

\subsection{Hyperparameters and experimental setups} 

We used PPO, TRPO, DDPG, TD3, and A2C default parameters used in the original papers. Table \ref{tab:parameters} provides a list of hyperparameters, simulator inputs, and further explanation for the setup of the conducted experiments. All the implementations and datas can be fount at \url{https://github.com/LightningCrashers/DyFEn}.

\subsection{B. LN Data}
\subsubsection{Data source: } We used the LN snapshots gathered by \citet{rohrer19lnattack} as our main source of data. These snapshots are originally generated by running a specific command on an LN node instance; therefore, a typical LN member can create a snapshot of its own. Each raw snapshot contains the profile of the public payment channels, but we tend to extract only the following info of each channel: source ID, target ID, channel ID, capacity, base fee, fee rate, min htlc, and last update. We conducted our experiments on the 2021-4-12 snapshot\footnote{\url{https://git.tu-berlin.de/rohrer/discharged-pc-data/-/tree/master/snapshots}}.

Additionally, we sourced our merchants data from 1ML\footnote{\url{https://1ml.com/}}. This data contains a list of merchant's node IDs. As described in Appendix A, merchants are sampled more frequently as receivers of random transactions. This modeling decision is compatible with the fact that merchants tend to receive a larger number of daily transaction inflow in comparison to regular nodes. 

\subsubsection{Data preprocessing: } We aggregate all channels between two nodes into one channel. We set the capacity of the new channel to the sum of the capacities, and the fee rate and the base fee to the average of fee rates and base fees of original channels. The reason behind this is that running the path searching algorithms on multi-directional graphs tends to degrade the performance and speed of the simulator without adding further optionality to it. Moreover, considering a single channel between two nodes satisfies the goal of this paper.

We also set the balances of each peer to half of the capacity. Additionally, the LEViN simulator supports this option to initiate the balances randomly or set the balance of the main node manually. Note that, balances are not included in the snapshots for the sake of privacy.

\section{C. Additional Experiments}
\begin{figure}[h]
\centering
\includegraphics[width=8.5cm]{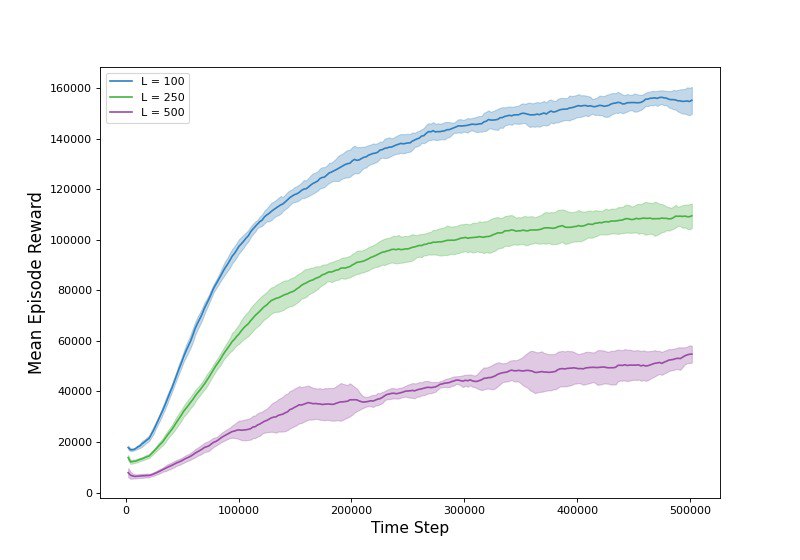}
\caption{\label{fig:local_train} Training curves of five runs for each of the three PPO models on node $a$. We simulated three different localization sizes $L=100,250,500$ with total of $30,75,150$ random transactions in every time step, respectively.}
\end{figure}

In this section, we provide the numerical results of the localization size impact on the convergence speed and the final performance of our best RL model in Figure \ref{fig:local_train} and Table 
\ref{tab:localization_table}. Table \ref{tab:static_results} summarizes the performance of the static strategy discussed in the paper using different approaches.

\begin{table}[h]
\begin{center}
\begin{tabular}{c c c} 
 \hline
 \textbf{Tested on} & \textbf{Trained on}  & \textbf{Discounted Income}  \\ [0.5ex] 
 \hline
  & $L=$100 &   149057.62  \\ 
  $L=$100 & $L=$250 &   114559.48  \\
  & $L=$500 &  92026.62\\
  \hline
  & $L=$100 & 72772.50  \\ 
  $L=$250 &  $L=$250 & 54181.78 \\
  & $L=$500 &  45409.50 \\
  \hline
  & $L=$100 &  41023.12 \\ 
  $L=$500 &  $L=$250 &  31525.63  \\
   & $L=$500 & 28011.54 \\
   \hline
   & $L=$100 &32861.84  \\ 
  $L=$1000 & $L=$250 & 24771.39 \\
  & $L=$500 &  28368.18  \\
 \hline
\end{tabular}
\caption{\label{tab:localization_table}PPO models trained on DyFEn with localization sizes $L=100,250,500$ on node $a$, each tested on four different localization sizes $L=100,250,500,1000$ averaged over five different random seeds.}
\end{center}
\end{table}

\begin{table*}[b]
\begin{center}
\begin{tabular}{p{3.5cm} c c c c c} 
 \hline
 \textbf{Policy} &\textbf{Node}& \textbf{Fee Rate} & \textbf{Base Fee}  & \textbf{Avg. Income} & \textbf{Std. Income}   \\ [0.5ex] 
 \hline
                &$a$&64065.05 &277928.18 &0 &0   \\ 
  \emph{Network avg.}  &$b$&64065.05 &277928.18 &0 &0  \\ 
                &$c$&64065.05 &277928.18 &0 &0   \\ 
 \hline
                     &$a$& 541.62&1760.82&0 &0    \\ 
  \emph{Local network avg.} &$b$& 337.66&1163.52 & 1457.43 & 434.65  \\ 
                     &$c$&195.21 &1052.48 &3588.05 &355.38   \\ 
 \hline
                &$a$& 50&1000 &1354.69 &538.41   \\ 
 \emph{ Network median } &$b$&50 &1000 &3283.79 &442.60   \\ 
                &$c$& 50 &1000 &2928.85 &460.71  \\ 
 \hline
                     &$a$& 125&1000 &0 &0    \\ 
  \emph{Local network median} &$b$&1 &1000 &4209.39 &667.63    \\ 
                     &$c$& 1&1000 &4207.20 &361.93  \\ 
 \hline
                 &$a$&1 &1000 &9814.04 &1320.17   \\ 
  \emph{Network mode}  &$b$ &1 &1000 &4209.39 &667.63  \\ 
                &$c$ &1 &1000 &4207.20 &361.93 \\ 
 \hline
                 &$a$&1 &1000 &9814.04 &1320.17   \\ 
 \emph{ Local network mode } &$b$ &1 &1000 &4209.39 &667.63  \\ 
                &$c$ &1 &1000 &4207.20 &361.93  \\ 
 \hline
\end{tabular}

\caption{\label{tab:static_results}Average income of nodes $a$, $b$, and $c$ using static strategies with different fee rates and base fees, showing the zero income for large fee rates and base fees.}
\end{center}
\end{table*}

\begin{table*}[b]
\begin{center}
\begin{tabular}{p{4cm} p{7cm} c}
 \hline
 \textbf{Name} & \textbf{Description}  & \textbf{Default}  \\ [0.5ex] 
 \hline
 \emph{Node Index} & Index of the center node in the LN snapshot. Agents interacting with DyFEn will try to maximize the income of this center node. &76620 \\ \\
 
 \emph{Base Fee Upper Bound} & Upper bound for base fee in the action space. The agent will choose strategies only with a base fee lower than this parameter. &10000 \\ \\
 
 \emph{Fee Rate Upper Bound} & Upper bound for fee rates in the action space. The agent will choose strategies only with a fee rate lower than this parameter. &1000 \\ \\

 \emph{Transaction Amounts} & A list containing the amounts of each different simulated transaction.   &(10000, 50000, 100000) \\ \\
 \emph{Transaction Counts} & A list containing the number of simulated transaction for each transaction amount. Length of this list should be the same as the transaction amounts parameter. &(10, 10, 10) \\ \\
 \emph{Epsilons} &A list containing the ratio of sampled transactions ending in a merchant.& (0.6, 0.6, 0,6) \\ \\
 \emph{Localization Size} & Size of the local network used in LEViN. Nodes of the local network will be chosen based on their distance from the center node. &100 \\ \\
 \hline
\end{tabular}

\caption{\label{tab:parameters}Description of the default parameters.}
\end{center}
\end{table*}

\end{document}